\documentclass{article} 
\usepackage{iclr2023_conference,times}

\usepackage{amsmath,amsfonts,bm}

\def\eqref#1{equation~\ref{#1}}

\def\1{\bm{1}}

\DeclareMathAlphabet{\mathsfit}{\encodingdefault}{\sfdefault}{m}{sl}
\SetMathAlphabet{\mathsfit}{bold}{\encodingdefault}{\sfdefault}{bx}{n}

\usepackage{hyperref}
\usepackage{lipsum}
\usepackage{url}
\usepackage{caption}
\usepackage{subcaption}
\usepackage{graphicx}
\newcommand{\figwidth}{0.3\textwidth}
\usepackage{booktabs}
\usepackage{multirow}
\usepackage{sidecap}
\title{Robustifying Deep Vision Models Through Shape Sensitization}

\author{
Aditay Tripathi 
\thanks{Work done at Google Research.}\\
Indian Institute of Science\\
\texttt{aditayt@iisc.ac.in}\\
\Andx
Rishubh Singh \\
Google Research\\
\texttt{rishubh@google.com} \\
\Andy
Anirban Chakraborty \\
Indian Institute of Science\\
\texttt{anirban@iisc.ac.in} \\
\Andz
Pradeep Shenoy \\
Google Research\\
\texttt{shenoypradeep@google.com} \\
}

\newcommand{\ouralgo}{{\sc ELeaS}}
\newcommand{\ouralgostr}{{\textbf{E}dge \textbf{Lea}rning for \textbf{S}hape sensitivity}}

     \makeatletter
    \def\footnoterule{\kern-3\p@
      \hrule \@width 2in \kern 2.6\p@} 
    \makeatother

\begin{document}

\maketitle

\begin{abstract}
Recent work has shown that deep vision models tend to be overly dependent on low-level or ``texture'' features, leading to poor generalization. Various data augmentation strategies have been proposed to overcome this so-called \textit{texture bias} in DNNs. We propose a simple, lightweight \textit{adversarial augmentation} technique that explicitly incentivizes the network to learn holistic shapes for accurate prediction in an object classification setting. Our augmentations superpose edgemaps from one image onto another image with shuffled patches, using a randomly determined mixing proportion, with the image label of the edgemap image. To classify these augmented images, the model needs to not only detect and focus on edges but distinguish between relevant and spurious edges. We show that our augmentations significantly improve classification accuracy and robustness measures on a range of datasets and neural architectures. As an example, for ViT-S, We obtain absolute gains on classification accuracy gains up to 6\%. We also obtain gains of up to 28\% and 8.5\% on natural adversarial and out-of-distribution datasets like ImageNet-A (for ViT-B) and ImageNet-R (for ViT-S), respectively. Analysis using a range of probe datasets shows substantially increased shape sensitivity in our trained models, explaining the observed improvement in robustness and classification accuracy.

\end{abstract}

\begin{figure}[h]
    \centering
    \includegraphics[width=\textwidth]{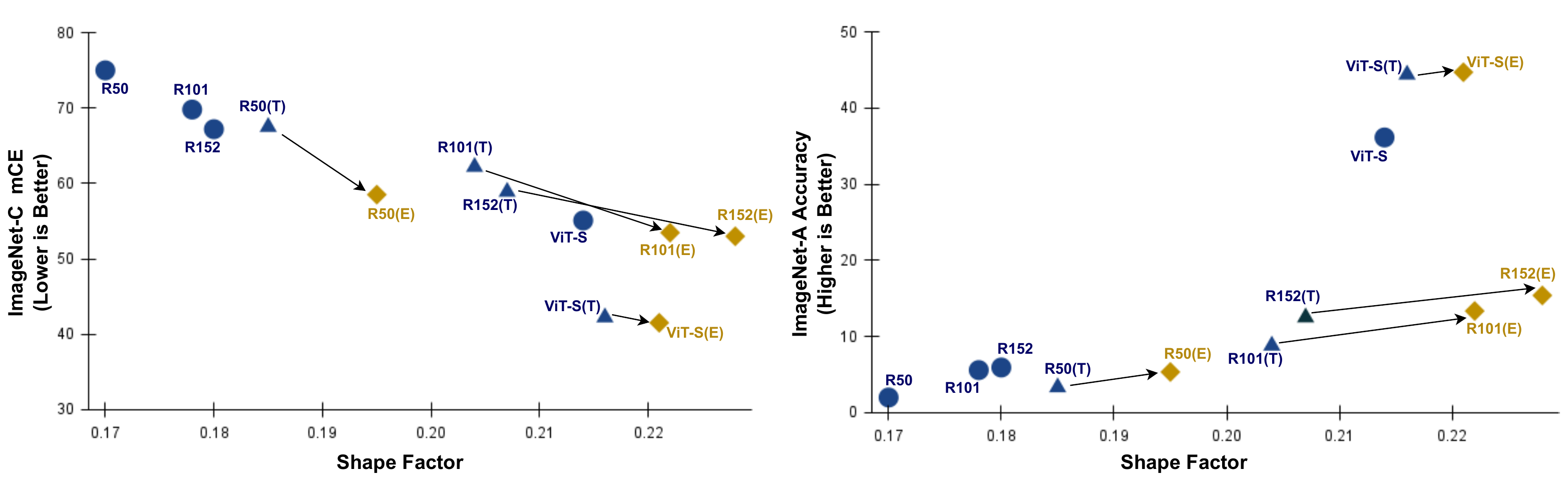}
    \caption{\textbf{Comparison of the models on robustness and shape-bias.} The shape factor gives the fraction of dimensions that encode shape cues~\cite{islam2021shape}. Backbone(T) denotes texture shape debiased (TSD) models~\cite{li2021shapetexture}. In comparison, \ouralgo\ denoted by Backbone(E) is more shape biased and shows better performance on ImageNet-C and ImageNet-A datasets.}
    \label{fig:teaser1}
\end{figure}

\section{Introduction}
A growing body of research catalogues and analyzes apparent failure modes of deep vision models. For instance, work on texture bias~\cite{geirhos2018imagenettrained,hendrycks2021nae,DBLP:journals/ploscb/BakerLEK18} suggests that image classifiers are overdependent on textural cues and fail against simple (adversarial) texture substitutions. Relatedly, the idea of simplicity bias~\cite{DBLP:conf/nips/ShahTR0N20} captures the tendency of deep models to use weakly predictive ``simple'' features such as color or texture, even in the presence of strongly predictive complex features. In psychology \& neuroscience, too, evidence suggests that deep networks focus more on ``local'' features rather than global features and differ from human behavior in related tasks~\cite{Jacob2021QualitativeSA}. More broadly speaking, there is a mismatch between the cognitive concepts and associated world knowledge implied by the category labels in image datasets such as Imagenet and the actual information content made available to a model via one-hot vectors encoding these labels. In the face of under-determined learning problems, we need to introduce inductive biases to guide the learning process. To this end, Geirhos et al.~\cite{geirhos2018imagenettrained} proposed a data augmentation method wherein the texture of an image was replaced with that of a painting through stylization. Follow-on work improved upon this approach by replacing textures from other objects (instead of paintings) and teaching the model to separately label the outer shape and the substituted texture according to their source image categories~\cite{li2021shapetexture}. Both these approaches discourage overdependence on textural features in the learned model; however, they do not explicitly incentivize shape recognition.

We propose a lightweight \textit{adversarial augmentation} technique \ouralgo\ (\ouralgostr) that is designed to increase shape sensitivity in vision models. Specifically, we augment a dataset with superpositions of random pairs of images from the dataset, where one image is processed to produce an edge map, and the other is modified by shuffling the location of image patches within the image. The two images are superposed using a randomly sampled relative mixing weight (similar to Mixup~\cite{DBLP:conf/iclr/ZhangCDL18}), and the new superposed image is assigned the label of the edgemap image. \ouralgo\ is designed to specifically incentivize not only edge detection but shape sensitivity: 1) classifying the edgemap image requires the model to extract and exploit edges -- the only features available in the image, 2) distinguishing the edgemap object category from the superposed shuffled image requires the model to distinguish the overall edgemap object shape (relevant edges) from the shuffled image edges (irrelevant edges, less likely to be ``shape like''). We perform extensive experiments over a range of model architectures, image classification datasets, and probe datasets, comparing \ouralgo\ against recent baselines. Figure~\ref{fig:teaser1} provides a small visual sample of our findings and results; across various models, a measure of \textit{shape sensitivity}~\cite{islam2021shape} correlates very strongly with measures of \textit{classifier robustness}~\cite{hendrycks2021nae,hendrycks2019robustness} (see Results for more details), validating the shape sensitivity inductive bias. In addition, for a number of model architectures, models trained with \ouralgo\ significantly improve both measures compared to the previous SOTA data augmentation approach~\cite{li2021shapetexture}.

Summing up, we make the following contributions:
\begin{itemize}
    \item We propose an adversarial augmentation technique, \ouralgo, designed to incentivize shape sensitivity in vision models. Our augmentation technique is lightweight (needing only an off-the-shelf edge detection method) compared to previous proposals that require expensive GAN-based image synthesis~\cite{geirhos2018imagenettrained,li2021shapetexture}. 
    \item In experiments, \ouralgo\ shows increased shape sensitivity on a wide range of tasks designed to probe this property. Consequently, we obtain increased accuracy in object classification with a 6\% improvement on ImageNet-1K classification accuracy for ViT-Small among others.
    \item \ouralgo\ shows high generalizability and out of distribution robustness with 14.2\% improvement in ImageNet-C classification performance and 5.89\% increase in shape-bias for Resnet152.
\end{itemize}

\begin{figure*}
    \centering
    \includegraphics[width=\textwidth]{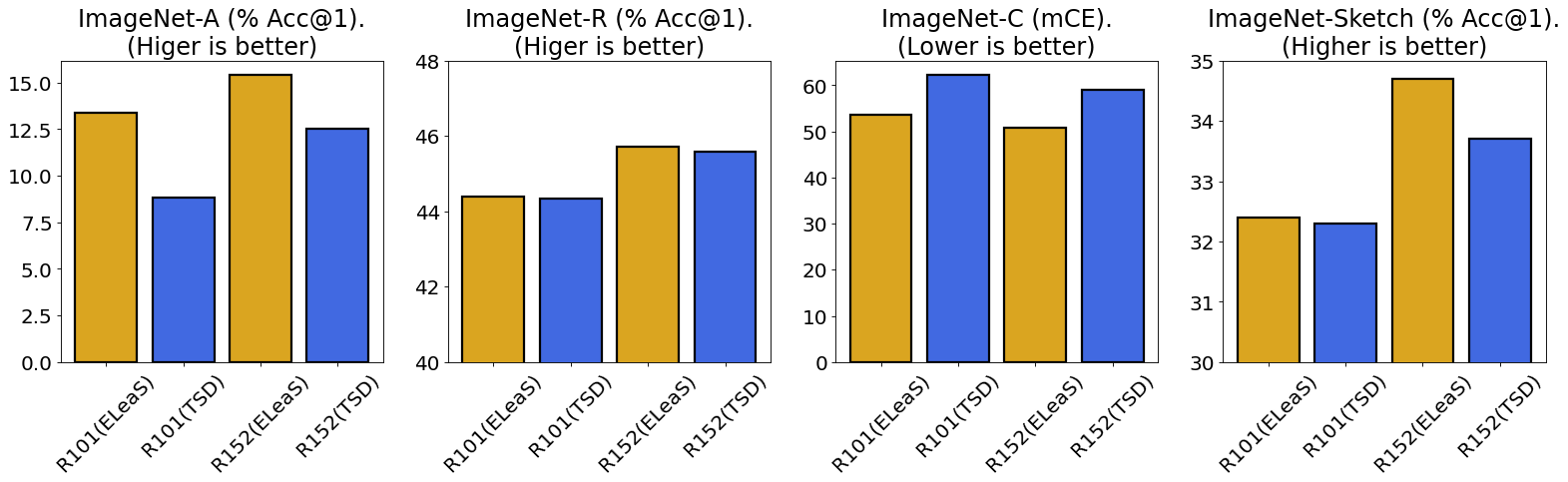}
    \caption{\textbf{Representative performance comparison} of our model, with the `Debiased' models (TSD\cite{li2021shapetexture}) on ImageNet-A, R, C and Sketch datasets. The models trained using  \ouralgo\ show improved performance on out-of-distribution robustness datasets. The large performance improvement on the ImageNet-A dataset indicates better robustness to natural adversarial examples. }
    \label{fig:teaser2}
\end{figure*}

\section{Related work}
\textbf{Texture bias in vision models.}
In~\cite{geirhos2018imagenettrained,DBLP:journals/ploscb/BakerLEK18}, convolutional neural networks are shown to be more sensitive towards the texture present in the image to classify the object correctly. Further, in order to mitigate the texture bias, they have suggested a training strategy where they have used modified images with random texture along with the natural images during training. However, utilizing images with conflicting textures leads to an unnatural shift in data distribution, leading to drops in performance on natural images. Instead, simple and naturalistic data augmentations strategies can also lead to an increase in shape bias of the CNNs without much loss in performance on natural images~\cite{hermann}. An increase in shape-bias in the CNN models is associated with an increase in the robustness of the models~\cite{geirhos2018imagenettrained,DBLP:conf/icml/ShiZDZMW20}, however~\cite{DBLP:conf/iclr/LiYTMT0YX21} suggested that a more balanced shape or texture biased models also leads to increase in the model robustness. The study of texture bias in vision models has been extended to the vision transformers (ViT)~\cite{DBLP:conf/iclr/DosovitskiyB0WZ21}, where the transformer models are found to be less texture biased than the convolutional models~\cite{Tuli2021AreCN}. 

A measure of shape/texture bias is proposed in~\cite{geirhos2018imagenettrained}, where they used images with conflicting textures obtained using style-transfer methods. However, this measure ignores a large portion of test images and is biased toward models trained using images modified by style-transfer methods. In~\cite{DBLP:conf/nips/HermannL20}, Hermann et al. proposed a linear classification layer in order to measure the shape decodability of the representations learned by the model. A more fine-grained shape evaluation measure was proposed in~\cite{DBLP:conf/iclr/IslamKEJODB21} where they compute the number of dimensions in the image representation that correlate with the shape features. Further, they proposed a shallow read-out module that takes feature representations from the model and predicts a segmentation map.  

\textbf{Data Augmentation for improving shape-bias}
In~\cite{geirhos2018imagenettrained}, authors suggested a data augmentation strategy in order to increase the shape bias in the trained models. They proposed the Stylized-ImageNet dataset, where they strip each image of its original texture and replace it with the style of a randomly selected painting through AdaIN style transfer~\cite{DBLP:conf/iccv/HuangB17}. During training, the model is trained to predict the category corresponding to the shape in the model, ignoring the texture, thus leading to an increase in the shape bias of the models. However, instead of changing the distribution of training images through style transfer, naturalistic data augmentation techniques such as  can also lead to improvement in the shape bias of the model~\cite{hermann}. 

\section{Methodology}
We now describe the training strategy used in \ouralgo\ to increase  the shape sensitivity of deep image classifiers.

Let $\mathcal{I}$, and $\mathcal{C}$ be the set of all images and their corresponding categories in the dataset. In standard classifier training, a classifier $\Theta$ is trained to predict the category $c \in \mathcal{C}$ of the image $i \in \mathcal{I}$. However, the image classifiers trained in this way are sensitive to the texture present in the natural images. In this work, we propose to use edge maps along with textures from natural images to increase  the shape sensitivity of  image classifiers.

\noindent \textbf{Obtaining shapes and textures:}
We approximate the shapes of objects by extracting edge-maps from images and natural textures by shuffling patches within images.  In particular, for each image $i \in \mathcal{I}$, an edge-map is constructed using an edge detection kernel (i.e., Laplacian kernel) to produce the set of all edge-maps or ``shapes'' $\mathcal{S}$. Similarly, a ``texture'' dataset $\mathcal{T}$ is generated by first dividing an image in $\mathcal{I}$ into 2$\times$2 patches and then randomly shuffling the patches within the image to obtain a patch-shuffled image.

\noindent \textbf{Generating augmentations:} We superimpose randomly selected pairs of images $s\in \mathcal{S}$ and $t\in \mathcal{T}$ to create new images, or augmentations, to add to the training dataset. Each such augmented image is assigned the label of the shape image $s$. The superimposed image is obtained  as follows:
\begin{equation}
    i_{s} = \lambda*t + (1-\lambda)*s
\end{equation}
where $\lambda$ is drawn randomly from a $\mathtt{Beta(\alpha, \beta)}$ distribution. The parameters $\alpha$ and $\beta$ are chosen such that a higher weight is given to the samples from set $T$. The random weighing parameter $\lambda$ introduces variation in the augmented image set, and helps to obtain better edge-map classification. Let's call the set of superimposed images $\mathcal{B}$.  

\noindent \textbf{Training procedure:}
Training proceeds via minibatch gradient descent, as is standard in current machine learning literature. We construct each mini-batch to have half of its images from the set of natural images $I\subset \mathcal{I}$ and the rest from the augmented dataset $B \subset \mathcal{B}$. During training, we minimize cross-entropy loss on natural image samples as well as our augmentations. In order to carefully control the degree of shape sensitivity induced by this process, we compute a weighted mixture of cross-entropy loss on these two image sets denoted here by $L$: 
\begin{equation}
    L(I,B,y_I, y_B) = \eta*CE(I,y_I) + 1-\eta* CE(B, y_B)
\end{equation}
where $\eta$ is varied to control the shape sensitivity, $CE$ is the cross-entropy loss, $y_I, y_B$ are the labels corresponding to the natural and augmented image sets $I,B$ respectively. To predict augmented images correctly, the model needs to interpret the edge-map present in the superimposed sample while at the same time ignoring distracting edges and textures from the other superimposed image. In this manner, we induce shape sensitivity in learned classifiers. The minibatch-mixing strategy encourages the model to generalize learned representations across natural and augmented images, thereby improving shape bias (and consequently overall accuracy) on natural images as well.

\section{Experimental setup}
\subsection{Models and training setup}
We train convolutional neural networks (CNNs) and Vision Transformers (ViTs) using our methodology. Among the CNN models, similar to \cite{li2021shapetexture}, we show results on ResNet50, ResNet101, and ResNet152. For training the ResNet models, we supplemented ImageNet data with an equal number of augmented images. We trained them for 100 epochs with a starting learning rate of $0.2$ which is reduced by a factor of 10 at the 30th, 60th and 90th epoch. Similar to TSD, while training the ResNet models, we also use auxiliary batch norm~\cite{DBLP:conf/cvpr/XieTGWYL20}. As ViTs are compute-intensive, we finetune ImageNet pretrained ViT models for 20k steps with a cosine learning rate schedule with a starting learning rate of $0.01$. The stochastic gradient descent (SGD) with a momentum of $0.9$ is used to train the models. We train all our models on 8 A100 GPUs with a batch size of 512 for ResNets, 256 for ViT-Small and ViT-Base and 128 for ViT-Large. We find that values $\alpha=4$, $\beta=1$, and $\eta=0.65$ produce the best results for all models. 

\subsection{Datasets and evaluation protocol}
\label{sec:protocol}

In this work, the models are trained on the publicly available ILSVRC 2012~\cite{ILSVRC15} dataset, which contains 1.28 million training images and 50k validation images. We evaluate and compare our trained models to answer the following questions:
\begin{itemize}
    \item Does \ouralgo\  lead to an increase in the shape-sensitivity of a model?
    \item Does the increased shape sensitivity result in better classification performance?
    \item Does the robustness of the models to distribution shift improve with increased shape sensitivity? 
\end{itemize}

The shape sensitivity of the models is first evaluated using the metric proposed in~\cite{geirhos2018imagenettrained}. The authors first created an image dataset called \textit{cue-conflict}, using a style-transfer method to transfer texture from one image to another. They then defined shape bias as the fraction of the shape decision when the model predicts either the shape or the
texture category. However, this metric ignores many a lot of images and is biased toward the methods that used Stylized-ImageNet during the training of the models. Further, authors in~\cite{islam2021shape} proposed a method to quantify the number of dimensions in the image representation that encode the object's shape. They take a pair of images with similar semantic concepts (i.e., shape) and then count the number of neurons that encode that specific concept. They have used Pascal VOC~\cite{Everingham10} and Stylized Pascal VOC dataset to create pair of images with particular semantic concepts. We have used these metrics to evaluate the shape bias in this work. Further, similar to~\cite{islam2021shape}, we also evaluate the quality of the shape encoding from these models by performing the binary and semantic segmentation on the frozen representations. PASCAL VOC is image segmentation dataset~\cite{Everingham10,DBLP:conf/iccv/HariharanABMM11}, with  10,582 training images and 1,449 validation images spanning across 20 object categories. 

The out-of-distribution generalization of the trained models is evaluated on four publicly available datasets, which are, ImageNet-A~\cite{hendrycks2021nae}, ImageNet-R~\cite{DBLP:conf/iccv/HendrycksBMKWDD21}, ImageNet-C~\cite{hendrycks2019robustness}, and ImageNet-Sketch~\cite{wang2019learning}. The ImageNet-A dataset consists of real-world natural adversarial images and is a challenging classification dataset. The ImageNet-R dataset consists of images with various renditions, such as cartoon art, DeviantArt, graphics, paintings, origami, etc., of the objects present in the original ImageNet dataset. The ImageNet-Sketch dataset consists of sketches of $1000$ ImageNet object categories. Both ImageNet-R and ImageNet-Sketch are used to evaluate the out-of-distribution generalization capability of the models. The ImageNet-C dataset consists of images with varying degrees of artificial distortions like `Gaussian Noise', `Motion Blur', `Speckle Noise', etc. Hence, it evaluates the robustness of the model to added distortions. 

\begin{table*}[!b]
\renewcommand*{\arraystretch}{1.1}
\centering
\begin{tabular}{l|lccccc}
\toprule
\multicolumn{1}{l|}{Model} & Method & IN-A($\uparrow$) & IN-R($\uparrow$) & IN-C($\downarrow$) & IN-Sketch($\uparrow$) & IN-1K($\uparrow$) \\
\midrule
\multirow{3}{*}{Resnet50}                  & Vanilla                 & 2.0        & 36.2      & 75.0       & 23.5      & 76.4    \\
                                           & TSD~\cite{li2021shapetexture}                & 3.3       & 40.8      & 67.5       & 28.3      & 76.9   \\
                                           & \ouralgo                    & \textbf{5.4}    & \textbf{41.7}    & \textbf{58.5}      & \textbf{29.7}      & \textbf{77.1}   \\
\midrule
\multirow{3}{*}{Resnet101}                 & Vanilla                 & 5.6        & 39.3       & 69.8       & 27.1      & 78.0   \\
                                           & TSD~\cite{li2021shapetexture}                & 8.8       & 44.3      & 62.2       & 32.3      & 7\textbf{8.8}    \\
                                           & \ouralgo                    & \textbf{13.4}      & \textbf{44.4}     & \textbf{53.5}     & \textbf{32.4}    & 78.6 \\
\midrule
\multirow{3}{*}{Resnet152}                 & Vanilla                 & 5.9       & 41.3      & 67.2       & 28.4      & 78.6   \\
                                           & TSD~\cite{li2021shapetexture}                & 12.5      & 45.5      & 58.9       & 33.3      & 79.7  \\
                                           & \ouralgo                    & \textbf{15.4}    & \textbf{45.7}     & \textbf{53.0}      & \textbf{34.7}      & 79.0  \\
\midrule
\multirow{3}{*}{ViT-S}                     & Vanilla                 & 16.6      & 36.1      &     55.1       &    33.2   & 74.6   \\
                                           & TSD~\cite{li2021shapetexture}                &  27.4          & 44.4           &  42.2          &  32.4     & 76.4       \\
                                           & \ouralgo                    & \textbf{28.3}     & \textbf{44.7}      &    \textbf{41.5}        &     \textbf{34.7}    &     \textbf{80.6} \\

\midrule
\multirow{2}{*}{ViT-B}                     & Vanilla                 & 34.6      & 40.7      &    50.8        &  45.8     & 79.5   \\
                                           & \ouralgo                    & \textbf{62.9}     & \textbf{56.4}      &     \textbf{34.0}       &  \textbf{46.4}       &     \textbf{85.5} \\

\midrule
\multirow{2}{*}{ViT-L}                     & Vanilla                 & 63.4      &    63.3   &      33.1      &   52.7    &  85.8  \\
                                           & \ouralgo                    & \textbf{67.4}     &  \textbf{65.9}     &     \textbf{29.5}       &      \textbf{54.1}   &     \textbf{86.2} \\
     
\bottomrule
\end{tabular}
\caption{\label{tab:IN} \textbf{Performance comparison on the ImageNet and the robustness datasets.} The models trained using \ouralgo\ show an improvement in the ImageNet performance along with better robustness. Except for IN-C the performance is measured in Accuracy@1 (\textbf{higher is better}). For IN-C, the performance is measured in mean corruption error (mCE) (\textbf{lower is better}). (\textbf{Refer to Sections~\ref{sec:class} and~\ref{sec:ood}})}
\end{table*}

Apart from ImageNet-C, the performance on these datasets is evaluated using classification accuracy. For the ImageNet-C dataset, the performance is evaluated using mean corruption error (mCE). We further evaluated the models on the out-of-distribution benchmark proposed in~\cite{geirhos2021partial}. 

\subsection{Baselines}
We have compared the performance of \ouralgo\ with the Shape-texture debiased(TSD) trained models~\cite{li2021shapetexture}. In TSD, the models are trained to focus on the image's shape as well as texture cues. It utilizes stylized ImageNet images containing conflicting shape and texture cues, and then the models are trained using the supervision from both the semantic cues. For TSD, among ViTs, only ViT-S is trained because of the large computation overhead involved in training ViT-B and ViT-L. \ouralgo\, on the other hand, learns the overall object shape in the presence of conflicting, less `shape-like' edges leading to increased shape sensitivity in the trained models.

\section{Results and discussion}
Our experiments study the impact of \ouralgo\ in the following stages: we show classification accuracy gains on the ImageNet dataset; we measure the robustness of trained models to a range of prediction challenges; and we examine \ouralgo's influence on shape-sensitivity of trained models and its affect on performance.

\subsection{Classification}
\label{sec:class}

Along with the increase in shape bias, \ouralgo\ also improves classification performance on the ImageNet dataset, in many cases by very large margins (Refer Table~\ref{tab:IN}). For instance, we see \textbf{$\mathbf{5.96\%}$ and $\mathbf{5.93\%}$ absolute improvement} in ImageNet classification accuracy for ViT-S and ViT-B respectively, compared to vanilla baseline, and a smaller but still significant 4.22\% gain in accuracy compared to TSD on VIT-S. We also see $0.6\%$, $0.62\%$, and $0.44\%$ increase in performance for ResNet50, ResNet101, and ResNet152 models respectively.

\begin{table*}[ht]
\renewcommand*{\arraystretch}{1.1}
\centering
\begin{tabular}{l|lccccc}
\toprule
Model                      & Method   & Edge  & Silhouette & Cue-conflict & Sketch & Stylized-IN \\
\midrule
\multirow{3}{*}{Resnet50}  & Vanilla  & 13.75 & 54.38      & 18.20        & 59.62  & 37.13       \\
                           & TSD\cite{li2021shapetexture} & 22.5  & \textbf{55.62}      & 21.40        & 67.0   & \textbf{56.13}       \\
                           & \ouralgo     & \textbf{35.62} & 54.37      & \textbf{21.41}        & \textbf{67.88}  & 46.0         \\
\midrule
\multirow{3}{*}{Resnet101} & Vanilla  & 23.85 & 49.37      & 19.92        & 63.12  & 41.75        \\
                           & TSD\cite{li2021shapetexture} & 31.25 & 51.87      & \textbf{24.92}        & 70.12  & \textbf{59.37}        \\
                           & \ouralgo     & \textbf{45.00} & \textbf{61.25}     & 24.06        & \textbf{71.88}  & 48.25        \\
\midrule
\multirow{3}{*}{Resnet152} & Vanilla  & 20.63 & 56.25      & 20.70        & 66.75  & 41.63        \\
                           & TSD\cite{li2021shapetexture} & 22.5  & \textbf{58.125}     & \textbf{25.31}        & 69.13  & \textbf{57.67}         \\
                           & \ouralgo     & \textbf{41.88} & 56.88      & 23.83        & \textbf{73.38}  & 48.75        \\
\midrule
\multirow{3}{*}{ViT-S} & Vanilla  & 25.0 & 26.25      & 22.96        & 49.12  & 44.25        \\
                           & TSD\cite{li2021shapetexture} & 22.5  & \textbf{46.87}     & \textbf{31.71}        & 68.62  & \textbf{77.5}       \\
                           & \ouralgo     & \textbf{34.38} & 43.13      & 27.66        & \textbf{69.63}  & 53.00       \\

\midrule
\multirow{2}{*}{ViT-B} & Vanilla  & 13.75 & 12.05      & 28.05        & 51.63  & 50.08        \\
                           & \ouralgo     & \textbf{41.25} & \textbf{66.25}      & \textbf{34.92}        & \textbf{82.00}  & \textbf{59.38}        \\

\midrule
\multirow{2}{*}{ViT-L} & Vanilla  & 55.62 & 68.75      & 40.39        & 83.75  & 65.50       \\
                           & \ouralgo     & \textbf{75.00} & \textbf{69.38}      & \textbf{42.11}        & \textbf{85.88}  & \textbf{66.50}       \\
\bottomrule
\end{tabular}
\caption{\label{tab:ood_hum} \textbf{Performance comparison on out-of-distribution benchmark proposed in~\cite{geirhos2021partial}.} The proposed method leads to significant improvement is performance over the vanilla Resnet models. The Cue-conflict and Stylized-IN utilizes stylized Images for evaluation and the Debiased method is trained on stylized images. Therefore the Debiased models show large performance on these datasets. (\textbf{Refer to Section~\ref{sec:ood_geirhos}})}
\end{table*}

\subsection{Out-of-distribution robustness}
\subsubsection{Evaluation on ImageNet-A, R, C, and Sketch datasets.}
\label{sec:ood}
As shown in Table~\ref{tab:IN}, \ouralgo\ significantly improves accuracy over both vanilla and TSD-trained models at these robustness challenges. For ImageNet-A, the considerable improvement (\textbf{4.54\%} vs. TSD on ResNet101) indicates better robustness to naturally occurring adversarial examples. Similarly, the  improvement in mCE on ImageNet-C  (\textbf{8.72\%} vs. TSD on ResNet101) showcases robustness to added distortions of various degrees. Similarly, \ouralgo\ shows robustness performance improvements on the ViT models. These large gains are driven primarily by \ouralgo's ability to better encode object shape (Refer Figure~\ref{fig:teaser1} for evidence supporting this causal link.)

\subsubsection{Evaluation on OOD benchmark~\cite{geirhos2021partial}}
\label{sec:ood_geirhos}

We also evaluated trained models on the OOD benchmark proposed in~\cite{geirhos2021partial}; results are in Table~\ref{tab:ood_hum}. The datasets in this benchmark consist of images from different domains such as `Edge', `Silhouette', `Sketch', `Stylized', and `Cue-conflict.', which require stronger shape sensitivity for correct classification. The `Cue-conflict' dataset is generated using iterative style transfer~\cite{DBLP:phd/dnb/Gatys17}, where the texture from a source image is transferred to a target image. Since these images contain cues corresponding to different objects, it is challenging to classify them without good shape-sensitive representations. Similarly, the Stylized imagenet (SIN) is generated by replacing the texture present in an image with style from a randomly selected painting using the AdaIN style transfer method~\cite{DBLP:conf/iccv/HuangB17}.  Models trained using \ouralgo\ can better encode object shapes in images, leading to large performance improvements on these datasets over the vanilla model. The better performance of TSD is expected as these models are trained using images from similar domains, i.e., the Stylized ImageNet dataset.

\begin{figure*}[ht]
\begin{subfigure}{\figwidth}
  \centering
  \includegraphics[width=\linewidth]{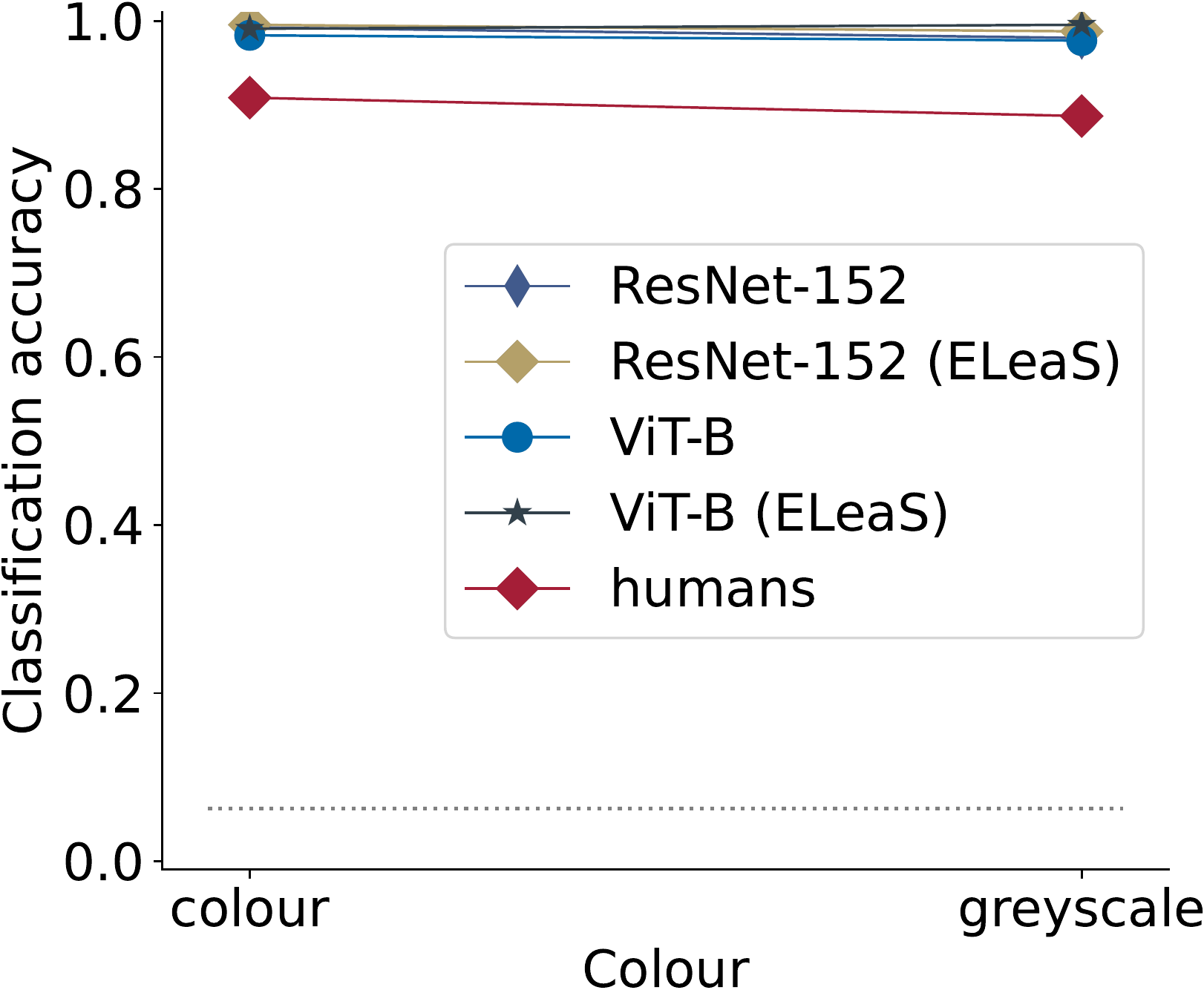}  
  \caption{True vs. false color}
  \label{fig:sub-first}
\end{subfigure}
\begin{subfigure}{\figwidth}
  \centering
  \includegraphics[width=\linewidth]{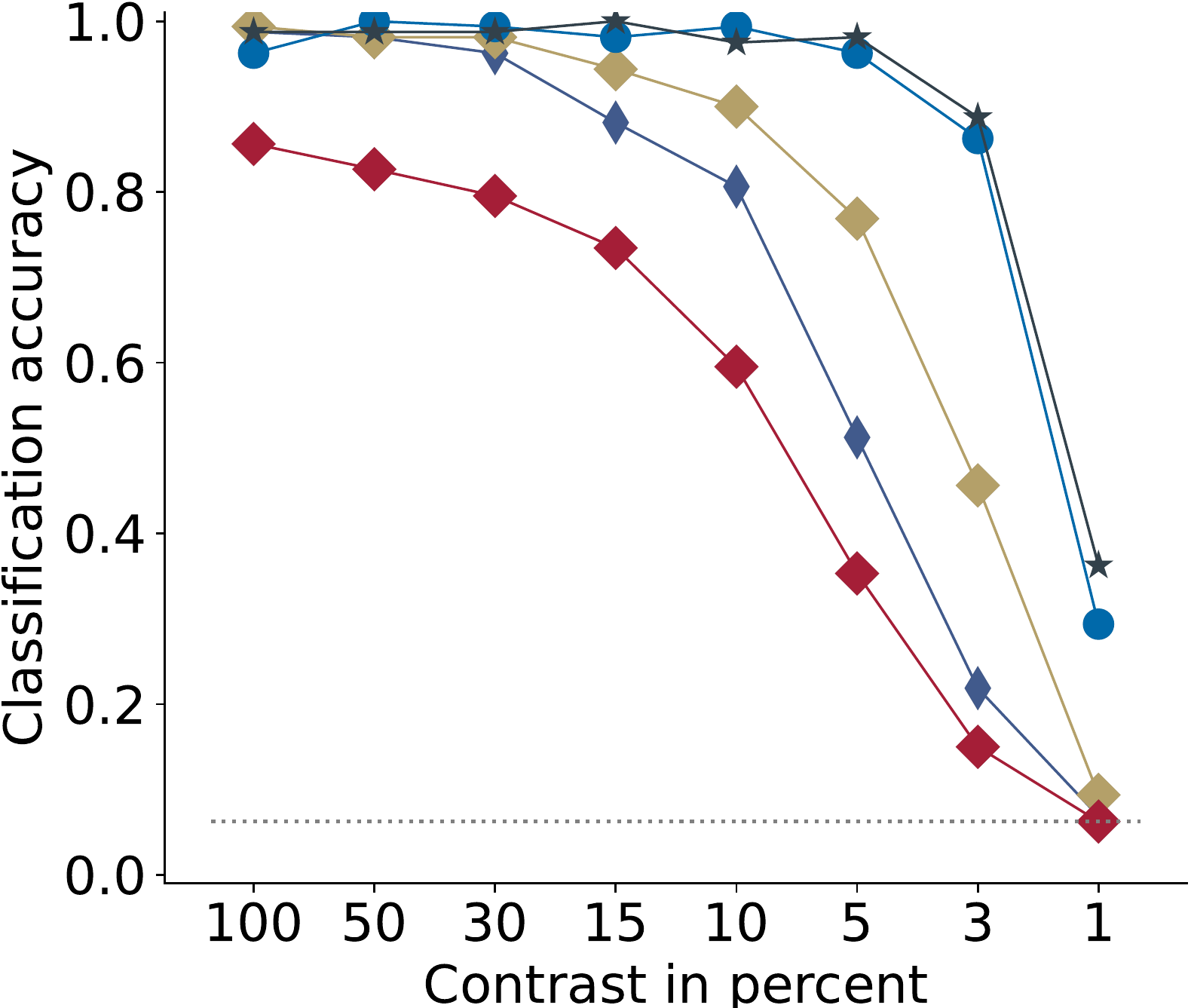}  
  \caption{Contrast}
  \label{fig:sub-second}
  \end{subfigure}
\begin{subfigure}{\figwidth}
  \centering
  \includegraphics[width=\linewidth]{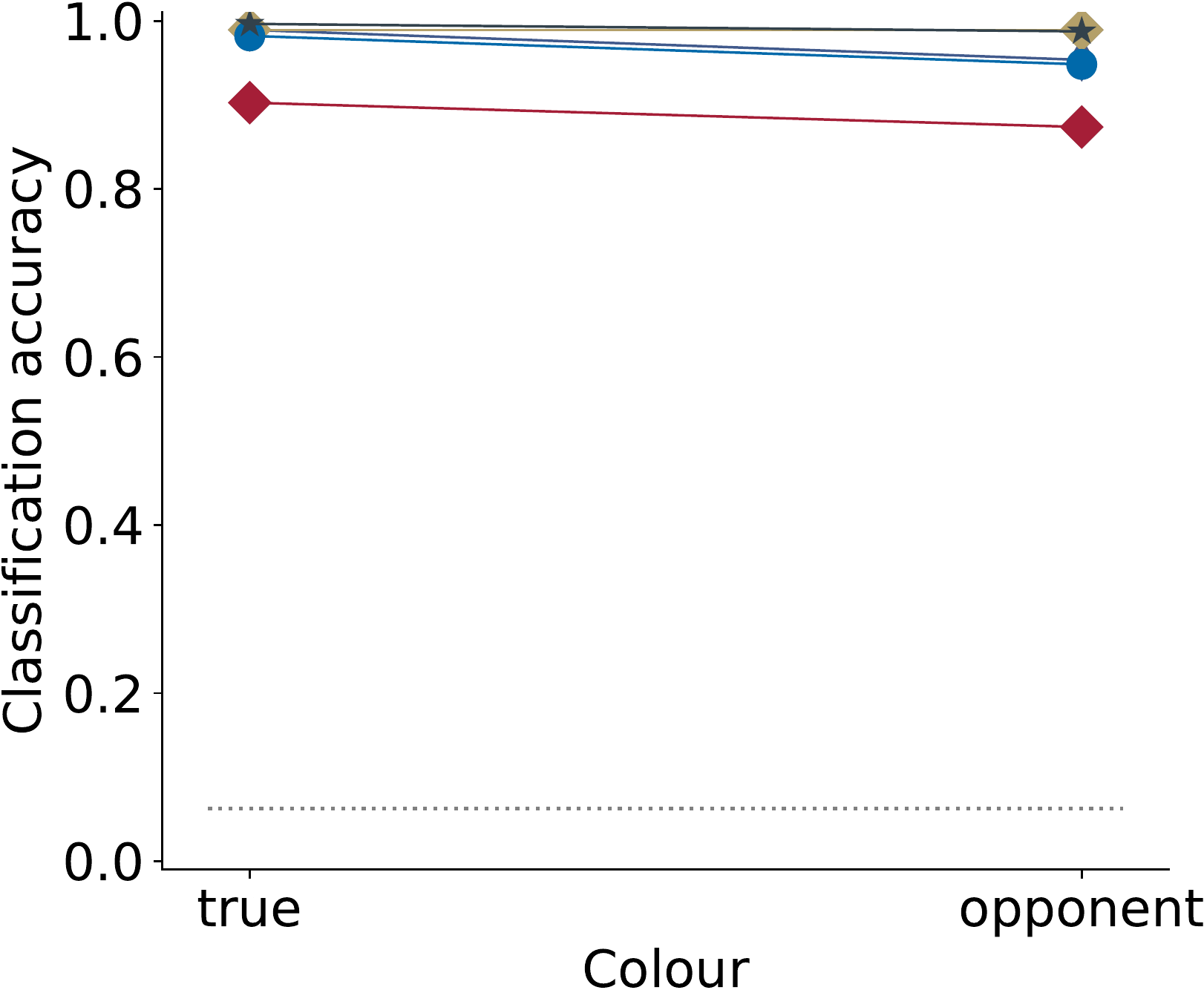}  
  \caption{Color vs greyscale}
  \label{fig:sub-third}
\end{subfigure}
\begin{subfigure}{\figwidth}
  \centering
  \includegraphics[width=\linewidth]{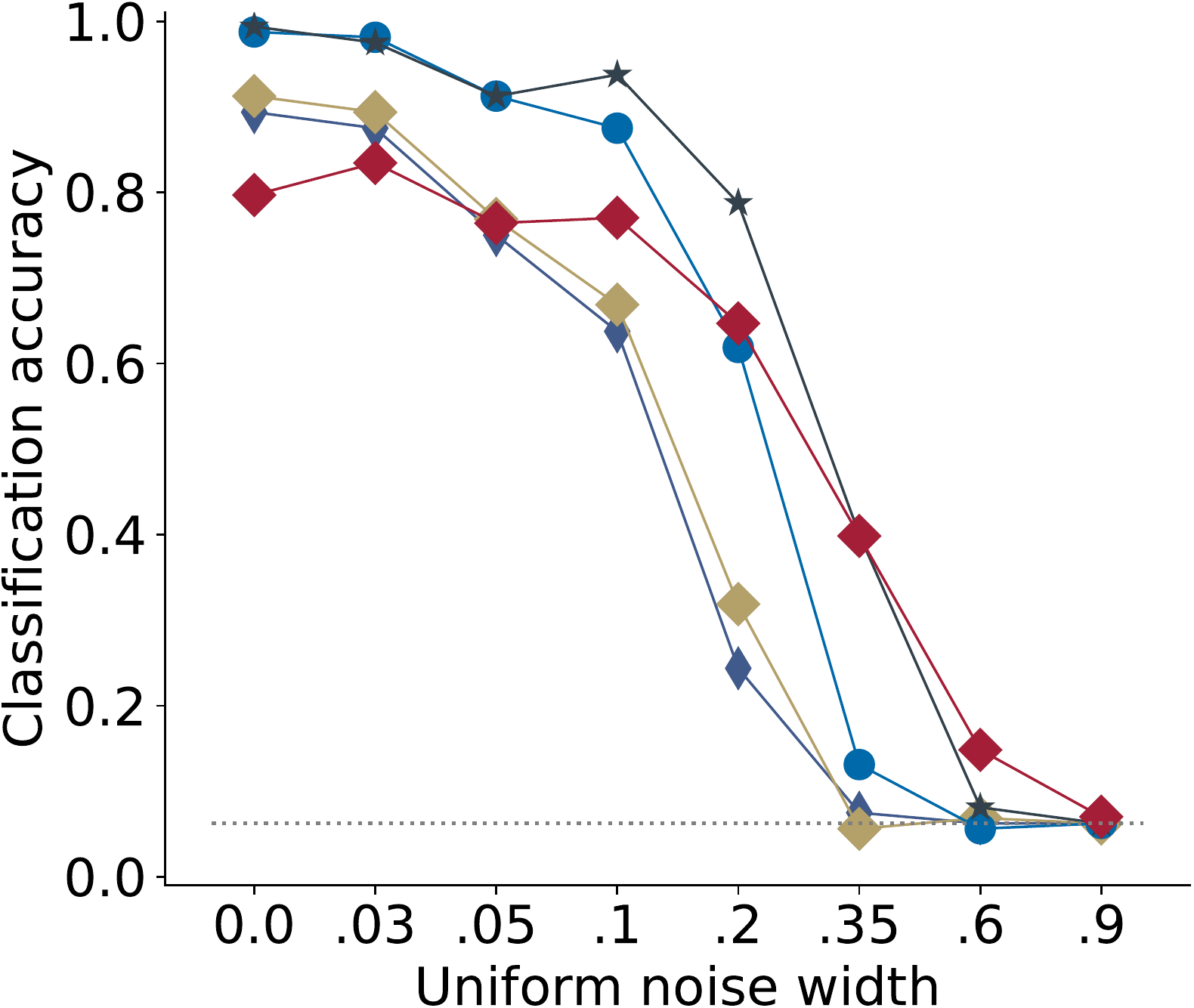}  
  \caption{Uniform Noise}
  \label{fig:sub-fourth}
\end{subfigure}
\begin{subfigure}{\figwidth}
  \centering
  \includegraphics[width=\linewidth]{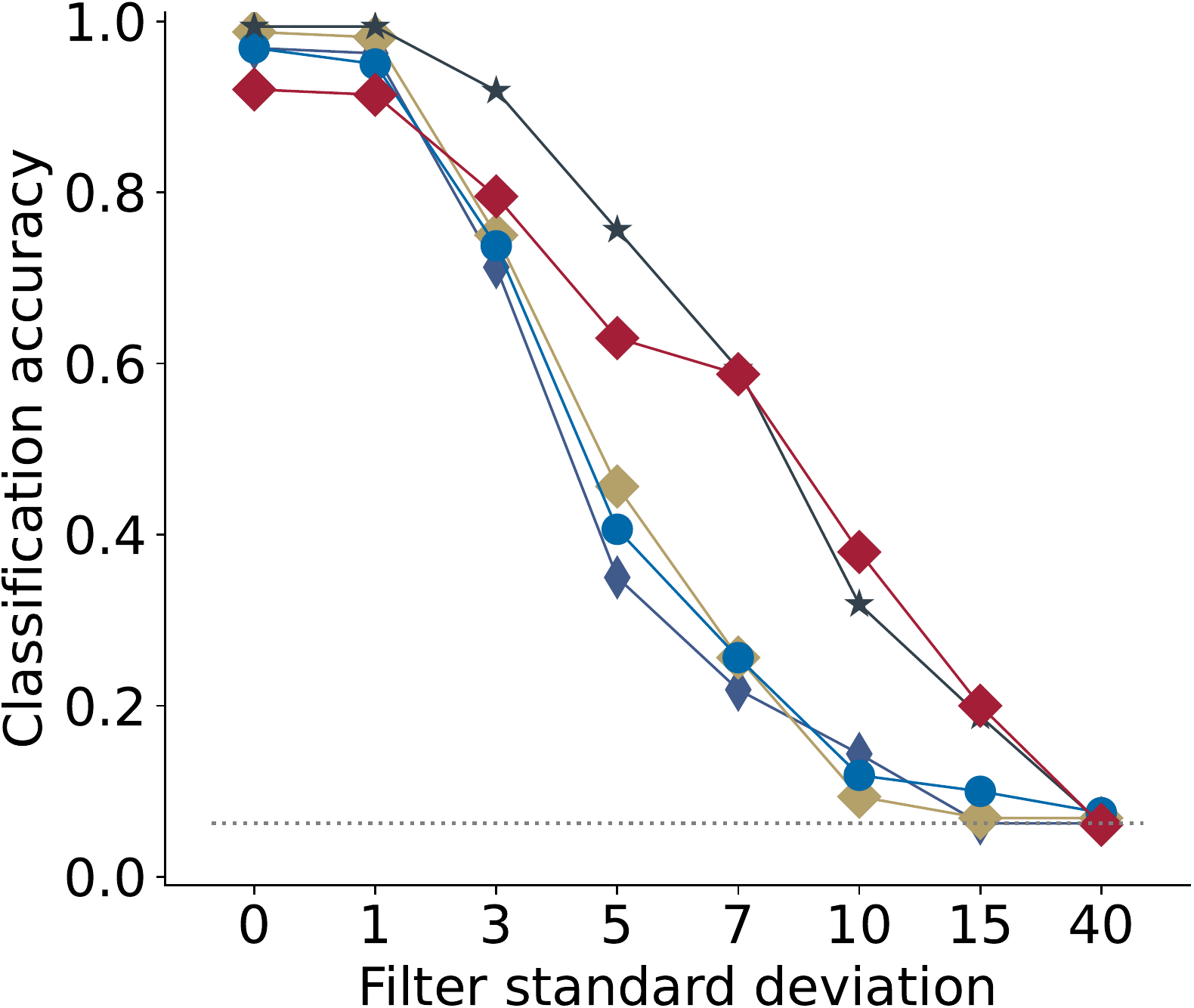}  
  \caption{Low-pass}
  \label{fig:sub-fifth}
\end{subfigure}
\begin{subfigure}{\figwidth}
  \centering
  \includegraphics[width=\linewidth]{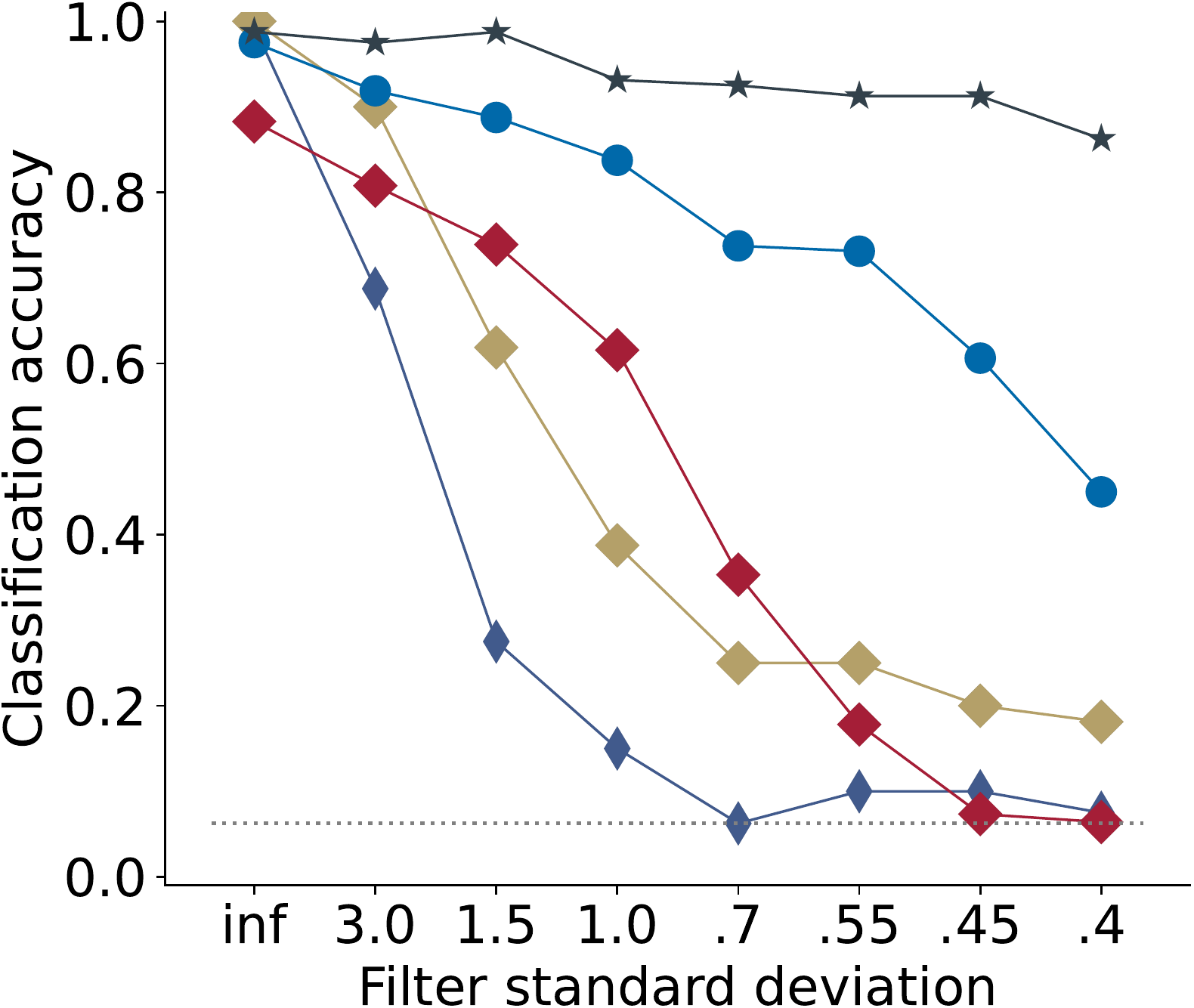}  
  \caption{High-pass}
  \label{fig:sub-sixth}
\end{subfigure}
\begin{subfigure}{\figwidth}
  \centering
  \includegraphics[width=\linewidth]{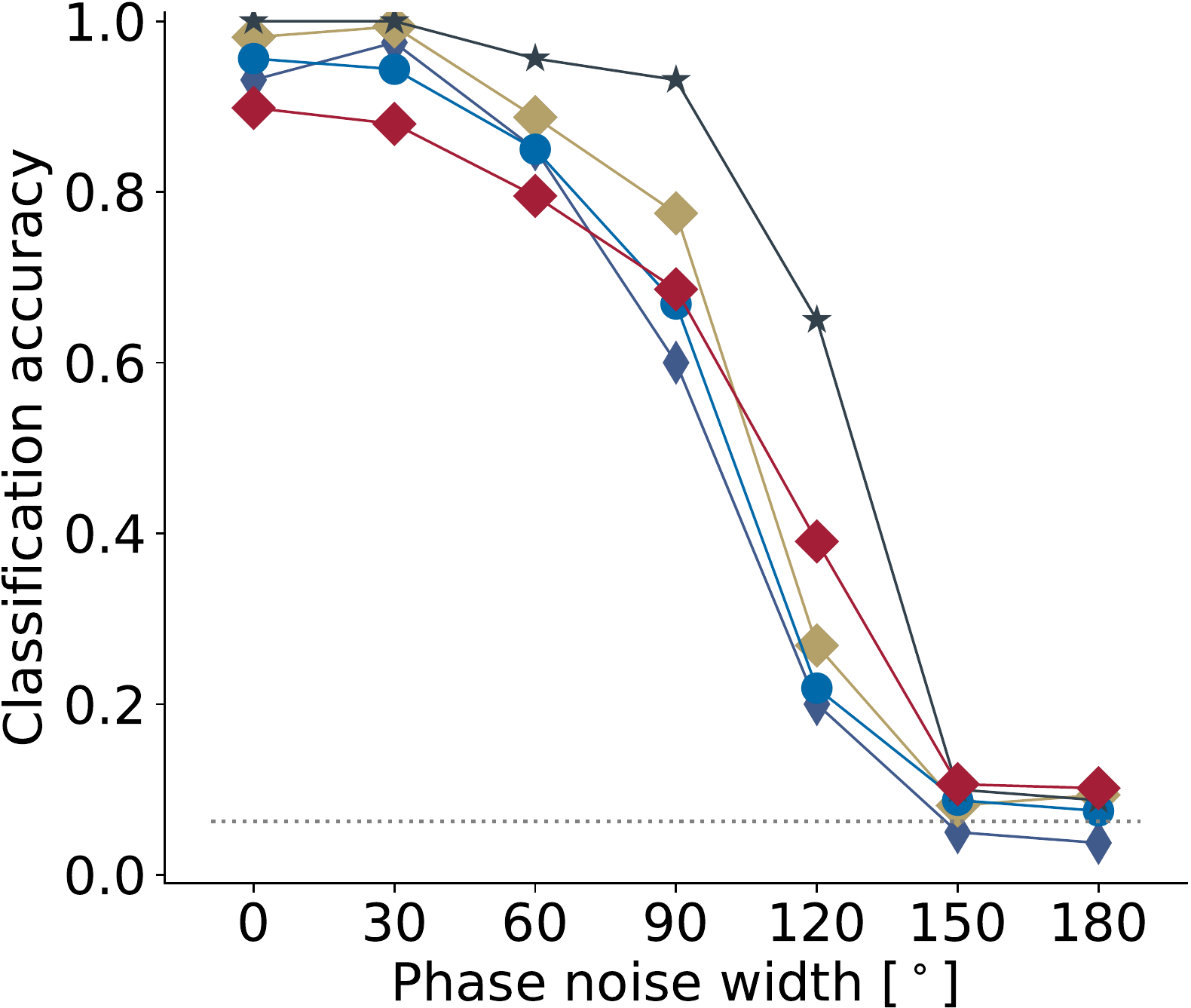}  
  \caption{Phase-noise}
  \label{fig:sub-seventh}
\end{subfigure}
\begin{subfigure}{\figwidth}
  \centering
  \includegraphics[width=\linewidth]{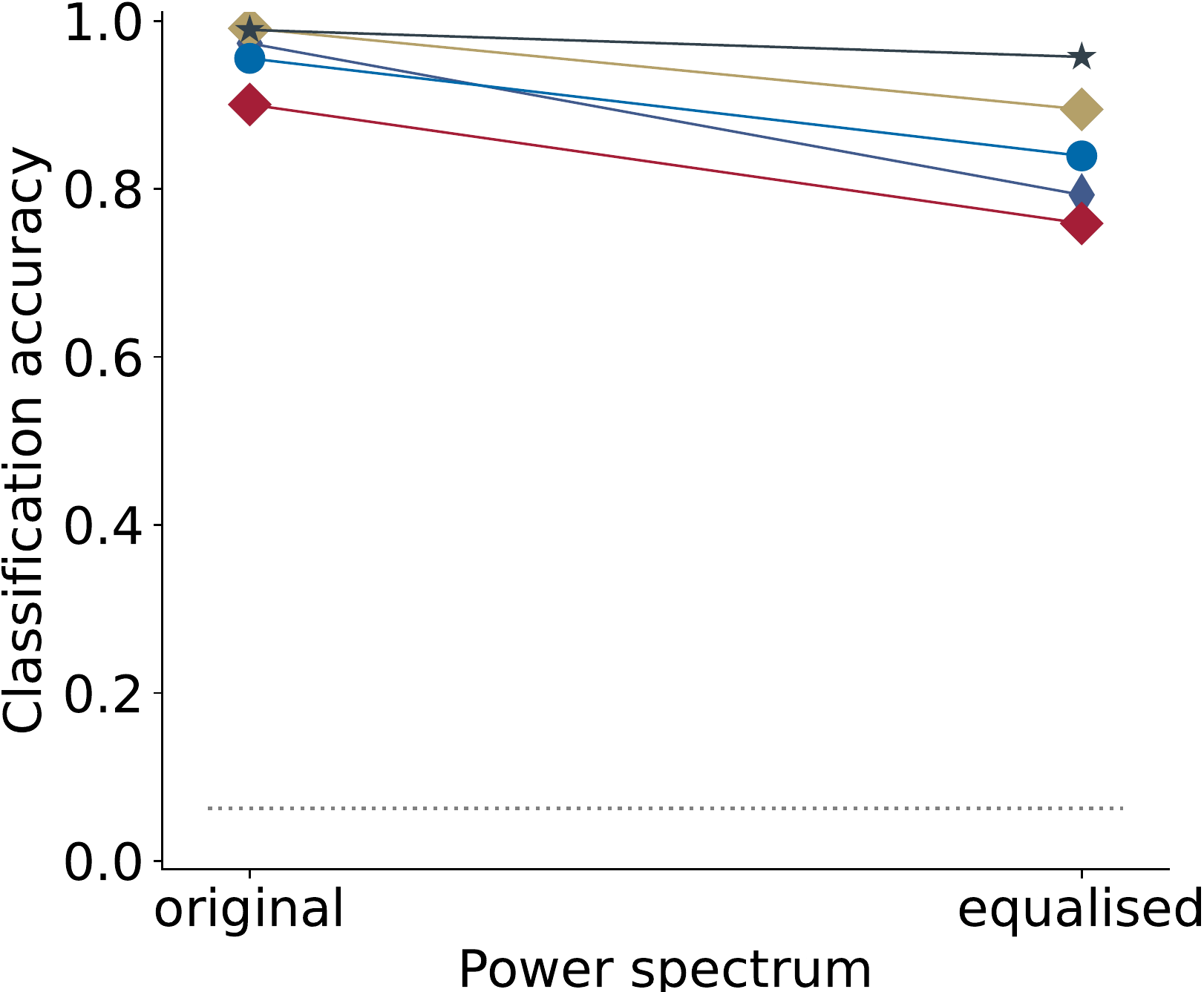}  
  \caption{Power Equalisation}
  \label{fig:sub-eighth}
\end{subfigure}
\begin{subfigure}{\figwidth}
  \centering
  \includegraphics[width=\linewidth]{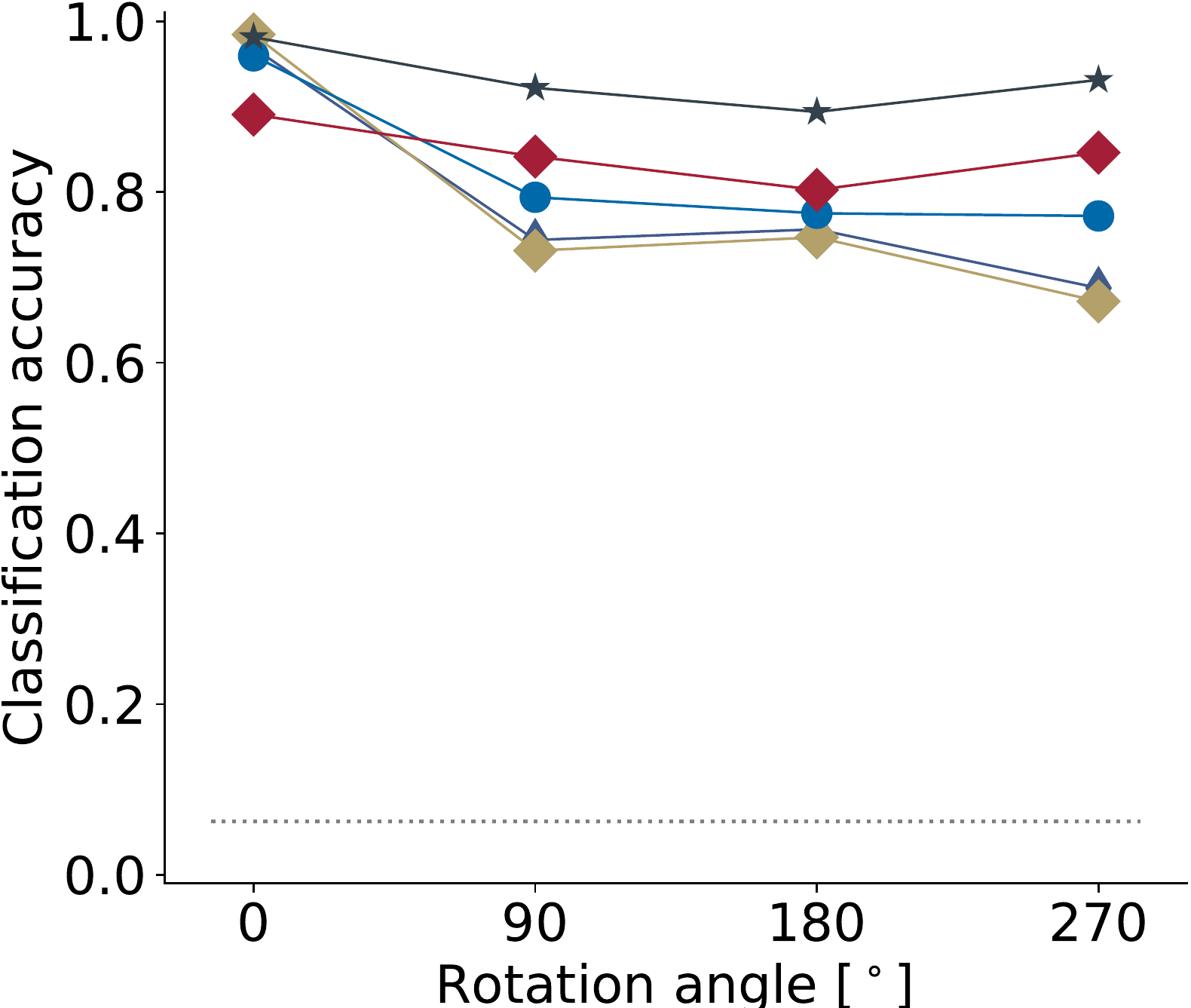}  
  \caption{Rotation}
  \label{fig:sub-ninth}
\end{subfigure}
\begin{subfigure}{\figwidth}
  \centering
  \includegraphics[width=\linewidth]{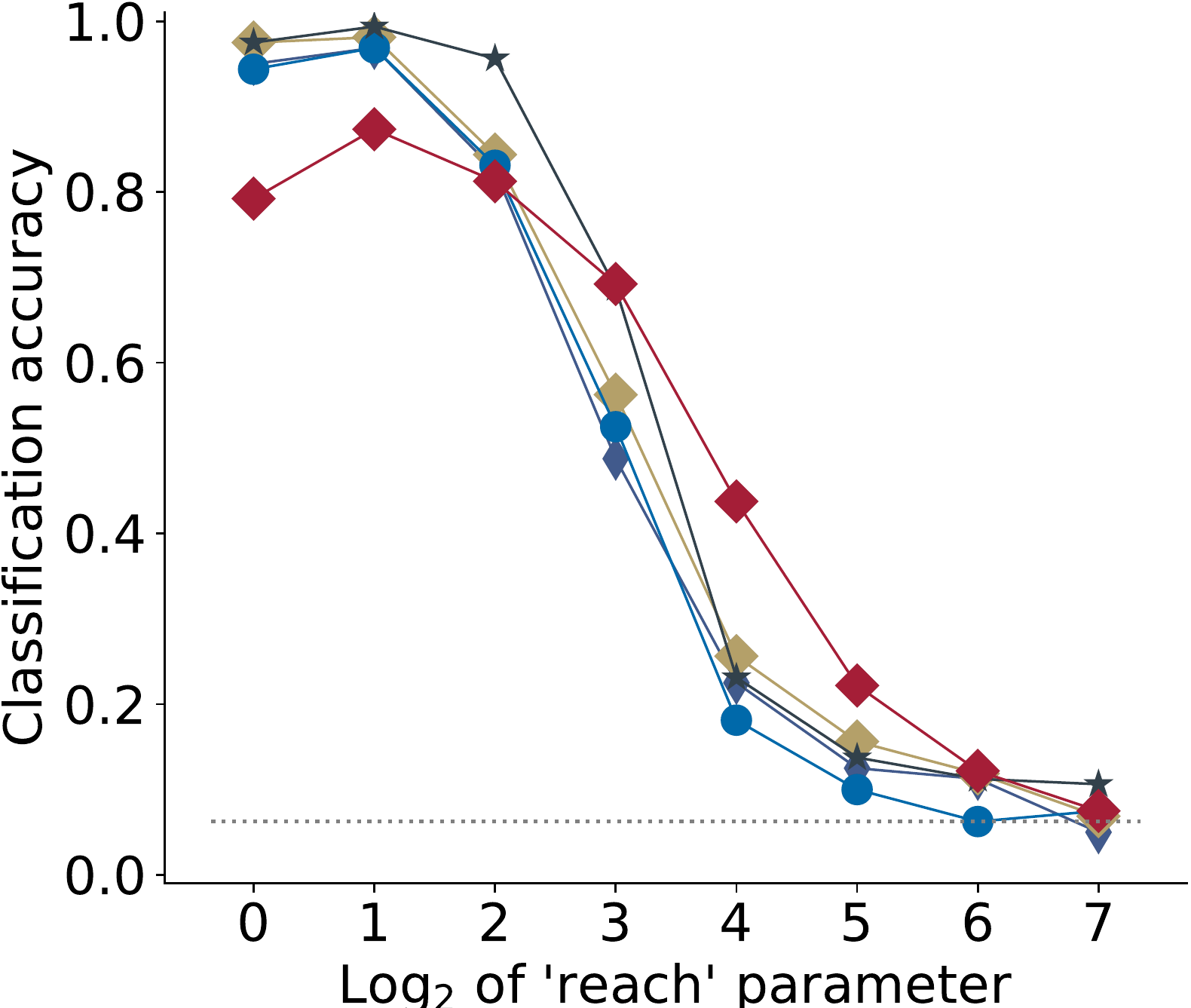}  
  \caption{EidolonI}
  \label{fig:sub-tenth}
\end{subfigure}
\hfill
\begin{subfigure}{\figwidth}
  \centering
  \includegraphics[width=\linewidth]{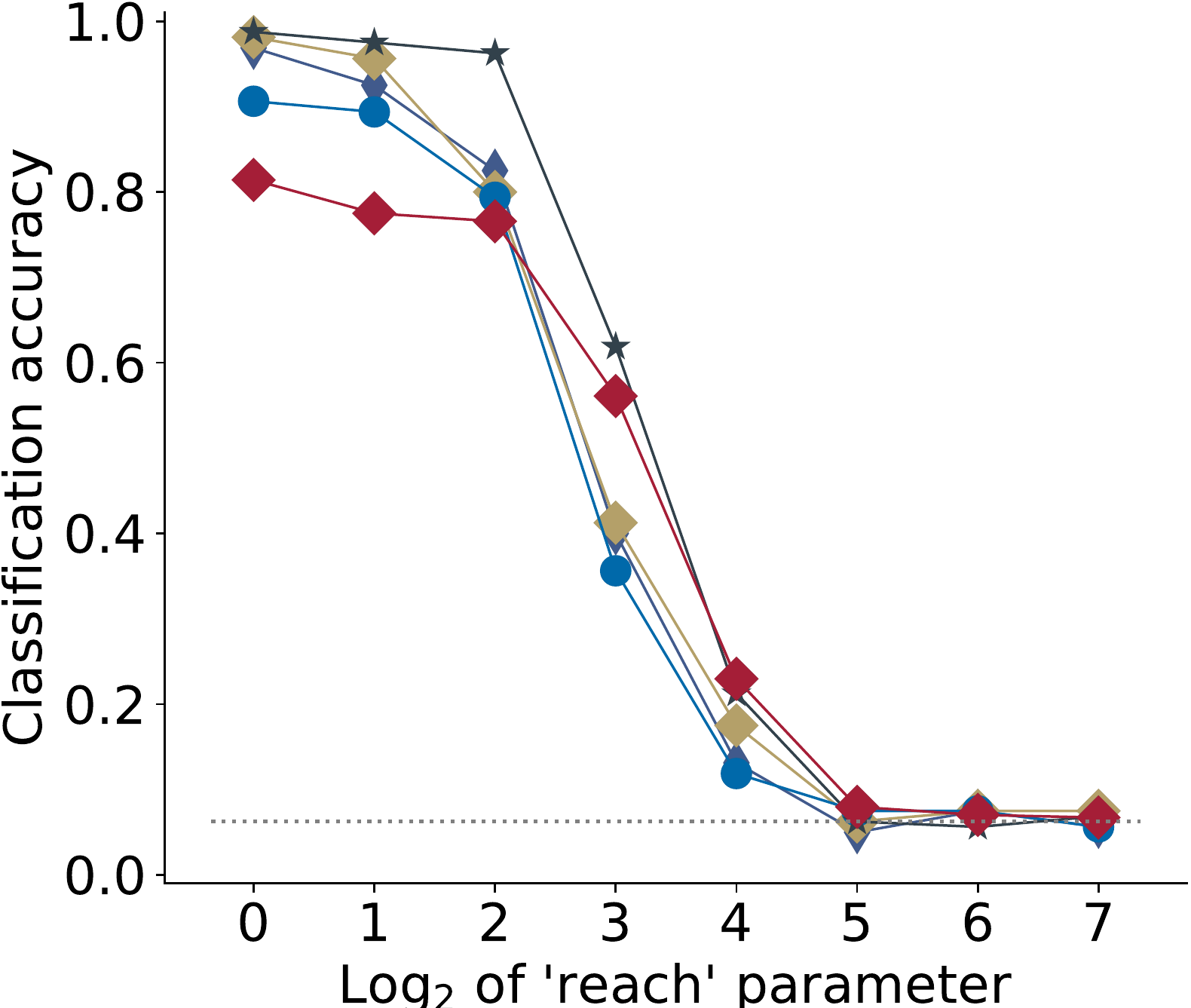}  
  \caption{EidolonII}
  \label{fig:sub-eleventh}
\end{subfigure}
\hfill
\begin{subfigure}{\figwidth}
  \centering
  \includegraphics[width=\linewidth]{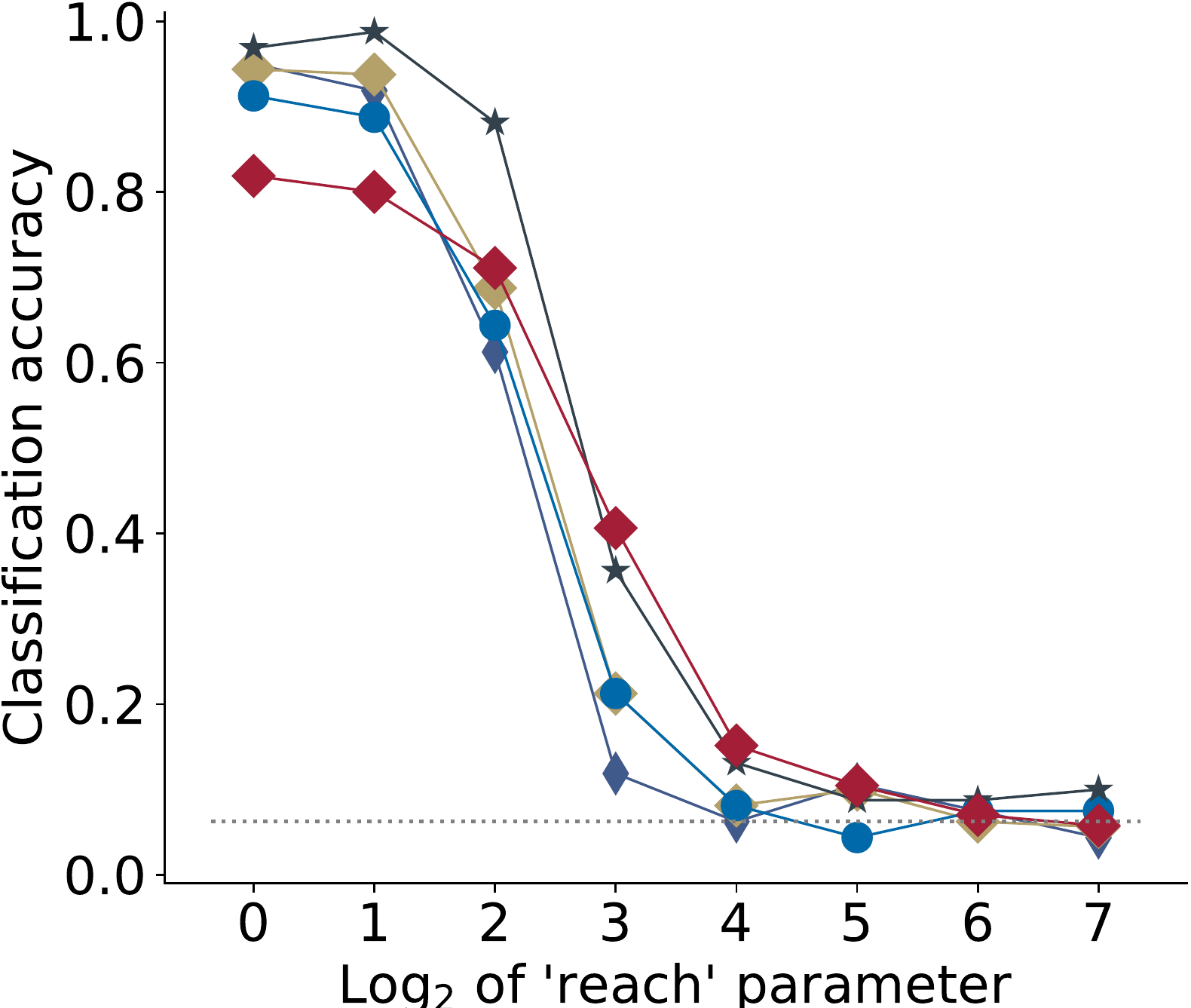}  
  \caption{EidolonIII}
  \label{fig:sub-twelveth}
\end{subfigure}
\caption{\label{fig:distortion}\textbf{Robustness comparison of \ouralgo\ trained Resnet152 and ViT-B with the human subjects.} \ouralgo\ leads to trained models which are more robust to added distortion and for many of the noise types it is more robust than the human subjects. (\textbf{Refer to Section~\ref{sec:ood}})}
\end{figure*}

We further evaluated robustness to distortion using the proposed benchmarks (Refer to Figure~\ref{fig:distortion}), which shows improved robustness to added distortions of varying degrees for \ouralgo.

\begin{figure}[!t]
    \centering
    \includegraphics[width=0.6\textwidth]{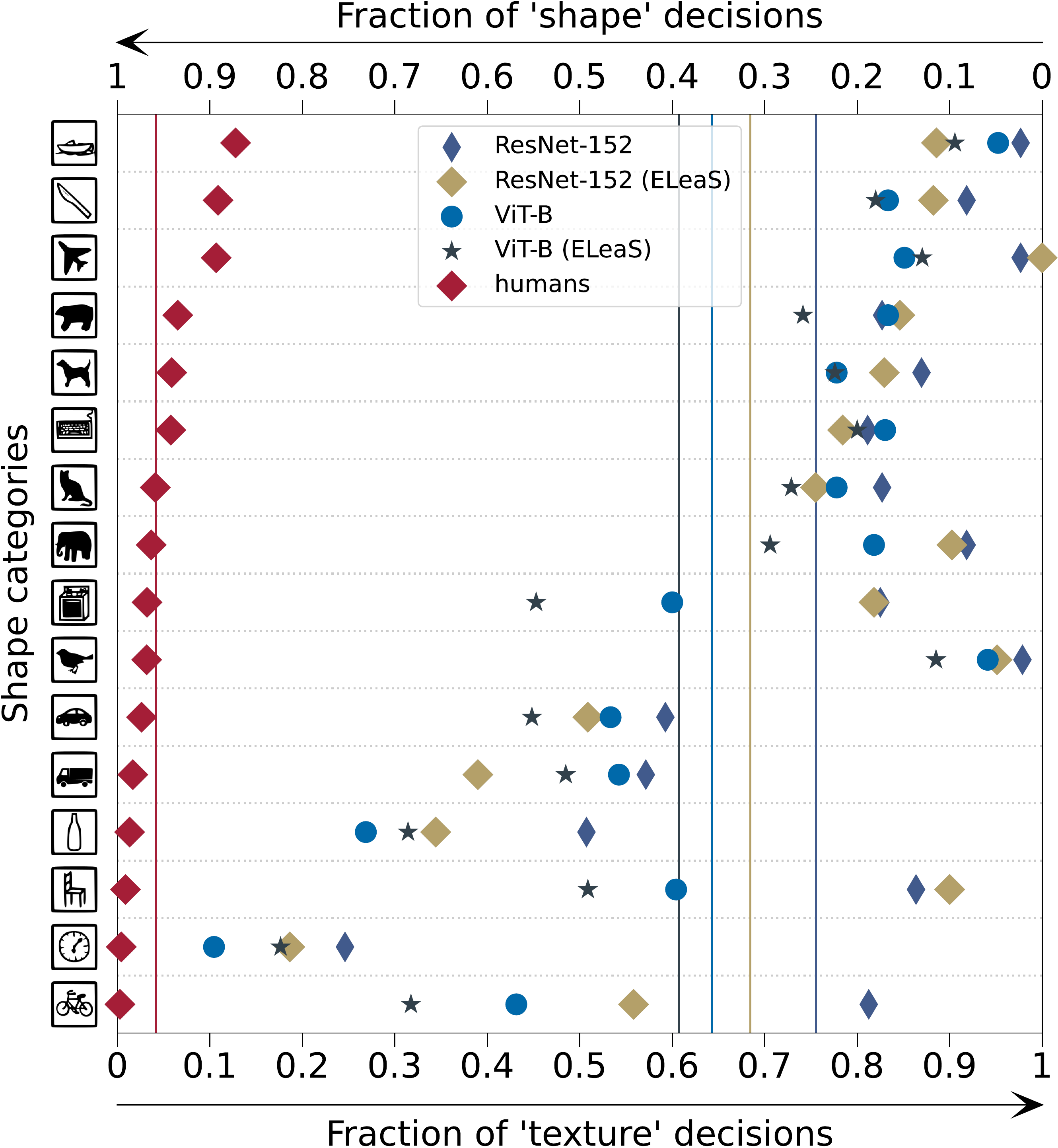}
    \caption{\protect\rule{0ex}{5ex}\textbf{Shape vs. texture bias for the vision models:} The images with conflicting shape and texture cues are used to calculate the shape bias of the trained models. The \ouralgo\ trained models show improvement in shape-bias. We have shown results for ResNet152 and ViT-L in this figure.(\textbf{Refer to Section~\ref{sec:shape_sense}})}
    \label{fig:shape-bias-resnet}
\end{figure}

\begin{table}[!b]
\renewcommand*{\arraystretch}{1.1}
\centering
\small
\begin{tabular}{l|lccc}
\toprule
\multirow{2}{*}{Model}     & \multirow{2}{*}{Method} & Shape- & \multicolumn{2}{c}{Factor $|z_k|$} \\
                          &                         &                bias             & Shape dim   & Texture dim   \\   
\midrule
\multirow{3}{*}{Res50}  & Vanilla                 & 19.75                       & 0.170         & 0.338       \\
                           & TSD\cite{li2021shapetexture}                & 23.11                       & 0.185         & 0.285            \\
                           & \ouralgo                    & 23.53                       & 0.195         & 0.299        \\
\midrule
\multirow{3}{*}{Res101} & Vanilla                 & 22.44                       & 0.178         & 0.323      \\
                           & TSD\cite{li2021shapetexture}                & 27.69                       & 0.204         & 0.256      \\
                           & \ouralgo                    & 26.04                       & 0.222         & 0.259     \\
\midrule
\multirow{3}{*}{Res152} & Vanilla                 & 21.74                       & 0.180         & 0.298       \\
                           & TSD\cite{li2021shapetexture}                & 28.78                       & 0.207         & 0.236         \\
                           & \ouralgo                    & 27.86                       & 0.228         & 0.249        \\   

\midrule
\multirow{3}{*}{ViT-S} & Vanilla                 & 33.33                       & 0.214         & 0.210      \\
                           & TSD\cite{li2021shapetexture}                & 36.56                       & 0.216         & 0.213           \\
                           & \ouralgo                    & 34.64                       & 0.221         & 0.217    \\   
\midrule
\multirow{2}{*}{ViT-B} & Vanilla                 & 33.12                       & 0.219         & 0.212     \\
                           & \ouralgo                    & 37.90                       & 0.239         & 0.205     \\   
\midrule
\multirow{2}{*}{ViT-L} & Vanilla                 & 47.04                       & 0.216         & 0.221         \\
                           & \ouralgo                    & 48.10                       &0.225 &0.207          \\       
                           \bottomrule
\end{tabular}
\caption{\label{tab:shape_bias} \textbf{Comparison of the shape-bias of the models.} The vision models trained using our strategy shows an increase in the fraction of `Shape' dimensions. However, the  shape-bias metric proposed by Geirhos et al.~\cite{geirhos2018imagenettrained} is biased towards the models trained using style-transfer datasets. (\textbf{Refer to Section~\ref{sec:shape_sense}})}
\end{table}

\subsection{Increasing shape sensitivity}
\label{sec:shape_sense}
We measure shape-sensitivity using the datasets and protocols described in Section~\ref{sec:protocol}; the results are in Tables~\ref{tab:shape_bias} and~\ref{tab:bin_sem_seg}.  Models trained using~\ouralgo\ show a large $6.12\%$ and $4.78\%$ increase in shape-bias for ResNet152 and ViT-L models, respectively, compared to vanilla baselines and are comparable to the shape-bias of models trained using TSD. \footnote{Note that shape bias~\cite{geirhos2018imagenettrained} has limitations: since it is defined in terms of a Stylized-ImageNet test set, the metric is biased towards methods that use SIN as a training augmentation. Further, it ignores images on which neither of the true shape/texture categories is predicted.} We further added class-wise shape bias comparison with human subjects in Figure~\ref{fig:shape-bias-resnet}.

An alternative measure, the shape factor~\cite{islam2021shape}, counts the number of dimensions in the image representation that encodes shape. \ouralgo\ substantially increases the number of shape dimensions in trained models (Table~\ref{tab:shape_bias}). We further evaluated shape decodability from learned representations by predicting the binary and semantic mask of the object in the image (Table~\ref{tab:bin_sem_seg}). The segmentation is performed by adding a three-layer readout module on top of the learned representations. The read-out module is trained on PASCAL VOC 2012 dataset~\cite{Everingham10}.

Resnet(TSD)~\cite{li2021shapetexture} shows a significant worsening of binary segmentation performance for all models and semantic segmentation performance for ResNet50, suggesting that their learned representations cannot predict pixel-wise object categories.  \ouralgo-trained model shows an increase in Semantic segmentation performance, indicating better decodability of pixel-wise semantic categories of objects.

\begin{table}[h]
\renewcommand*{\arraystretch}{1.25}
\centering
\begin{tabular}{l|lcc}
\toprule
Model     & Method &  Bin             & Sem      \\   
\midrule
\multirow{3}{*}{Resnet50}  & Vanilla       & 79.8            & 61.6           \\
                          & TSD\cite{li2021shapetexture}   & 78.9            & 61.1    \\
                          & \ouralgo        & 79.2            & 61.9           \\
\midrule
\multirow{3}{*}{Resnet101} & Vanilla                 & 80.4            & 63.4           \\
                          & TSD\cite{li2021shapetexture}   &  80.0        &     64.7       \\
                          & \ouralgo     & 81.0            & 65.9           \\
\midrule
\multirow{3}{*}{Resnet152} & Vanilla        &   80.3              &   64.4             \\
                          & TSD\cite{li2021shapetexture}       &     79.8     &  65.5\\
                          & \ouralgo  & 81.0             &      66.6     \\   

\bottomrule
\end{tabular}
\caption{\label{tab:bin_sem_seg} \textbf{Shape decodability of the learned representations:} The amount of shape information is evaluated by adding read-out modules on the frozen learned representation and then predicting the binary (Bin) and semantic (Sem) segmentation masks~\cite{islam2021shape}. The performance is reported in mIoU. The Pascal VOC dataset is used here to evaluate the shape decodability of the models. (\textbf{Refer to Section~\ref{sec:shape_sense}})}
\end{table}

\subsection{Lightweightness of \ouralgo\ }
\ouralgo\ is very lightweight compared to previous competing augmentation methods. The edge maps and shuffled patch images can be computed and stored on disk beforehand. To obtain the adversarial augmented image, computing a weighted combination of two images requires a few operations. In comparison, TSD\cite{li2021shapetexture} needs to run a GAN-based stylization to generate every augmented image making it much slower.

GAN-based stylized image generation takes almost double the time (keeping all other factors constant -- 8 A100 GPUs in our case) compared to that required for computing all edge maps and shuffled patch images. Both \ouralgo\ and TSD need to compute new augmented images with different edge map (source) images and shuffled patch (texture) images every epoch, so GAN-based augmentation must be run every epoch, making it expensive. Because of the aforementioned computational requirements, the per epoch cost of \ouralgo\ is negligible compared to TSD.

Moreover, for \ouralgo, once computed, the edge maps and shuffled patch images can be repeatedly used for training any models reducing the average data generation time cost even further.

\section{Conclusion}

We propose and evaluate \ouralgo\ -- a lightweight \textit{adversarial augmentation} technique for image classification in order to induce a shape bias in learned models. Previous work in this area~\cite{geirhos2018imagenettrained,li2021shapetexture} identified the need for such inductive biases due to the apparent dependence of deep models on textural features; however, their proposed augmentation techniques have drawbacks--first, they primarily sever the dependence on texture, while not explicitly enforcing processing and representation of shape, and second, the proposed augmentation process is very expensive. Although our work, too, proposes only data augmentation and does not come with guarantees on learned representations, the augmentations are simple to compute and are designed to encourage holistic shape representation. Extensive experimentation shows both the value of this inductive bias (shape factors correlate with model robustness and accuracy across models) and the substantial gains of our proposal, including over 5\% absolute accuracy improvements on ImageNet using vision transformers. An interesting direction for future work is to explore richer ways of specifying ``shape'' to the models--for instance, using off-the-shelf segmentation or depth estimation models--and also to explore other inductive biases that can profitably be incorporated into vision models. 

\bibliography{iclr2023_conference}
\bibliographystyle{iclr2023_conference}


\end{document}